\documentclass[aps]{revtex4}
\usepackage[latin1]{inputenc}
\usepackage{graphics}
\usepackage{times}
\usepackage[T1]{fontenc}
\usepackage[dvips]{graphicx}
\usepackage{amsmath,amsthm,amsfonts,amssymb}
\usepackage{float}

\begin{document}

\title{Obtaining Membership Functions from a Neuron
Fuzzy System extended by Kohonen Network}

 \author {Angelo Luís Pagliosa}
 \affiliation{Departamento de Engenharia Elétrica,
 Universidade do Estado de Santa Catarina,
  89223-100 Joinville, SC, Brazil}

 \author {Claudio Cesar de Sá}
 \email{claudi@joinville.udesc.br} \affiliation{Departamento de
 Ciências da Computação, Universidade do Estado de Santa Catarina,
  89223-100 Joinville, SC, Brazil}

 \author{F. D. Sasse}%
 \email{fsasse@joinville.udesc.br}
 \affiliation{Departamento de Matemática, \\Universidade do Estado de
 Santa Catarina  89223-100 Joinville, SC, Brasil}

 \begin{center}
\author {Angelo Luís Pagliosa \footnote{Departamento
de Engenharia Elétrica, Universidade do Estado de Santa Catarina -
CCT/UDESC, 89223-100 Joinville, SC, Brazil}, Claudio Cesar de
Sá\footnote{Departamento de Ciências da Computação, Universidade
do Estado de Santa Catarina - CCT/UDESC, 89223-100 Joinville, SC,
Brazil, claudio@joinville.udesc.br} and F. D.
Sasse\footnote{Departamento de Matemática, Universidade do Estado
de Santa Catarina - CCT/UDESC, 89223-100 Joinville, SC, Brazil,
fsasse@joinville.udesc.br}}
\end{center}
\begin{abstract}
This article presents the Neo-Fuzzy-Neuron
    Modified by Kohonen Network (NFN-MK), an hybrid computational model
    that combines fuzzy system techniques
    and artificial neural networks. Its main task consists in the automatic generation of
    membership functions, in particular, triangle forms, aiming a dynamic modeling of a
    system. The model is tested by simulating  real systems, here represented by a nonlinear
    mathematical function.  Comparison with the results obtained by
    traditional neural networks, and correlated studies of neurofuzzy systems applied in
    system identification area, shows that the  NFN-MK model has a similar
    performance, despite its greater simplicity.
\end{abstract}
 \maketitle

\section{Introduction}
     A traditional approach to  Artificial Intelligence (AI) is
     known as {\em connectionism},
     and its represented by    the field of  Artificial Neural
    Network (ANN). A second approach to AI is the {\em symbolic}
    one, with its various branches,  Fuzzy
    Logic (FL) among them.  ANNs models offer the possibility of  learning
    from input/output data and its functionality  is  inspired by biological neurons.
     Normally, ANNs   require a relative
    long training time and cannot be described as a mechanism
    capable to explain how its results were
    obtained by  training. Therefore, some
    projects involving ANNs can become complex, also lacking
      strong foundations.
    On the other hand, the FL framework deals with
    approximate reasoning typical from human
    minds. It allows the
    use of linguistic terminology  and common sense knowledge.
     The main limitation of this system
    consists in the nonexistence of learning
    mechanisms, capable of generating  fuzzy
    rules  and membership functions, which
    depend on the specialist knowledge.

    An interesting alternative consists of dealing with hybrid
    systems (HS), which use the advantages of both ANNs and FL.
    It can also employ the  process model  knowledge in order to
    decrease the duration of the project.
    In particular, HS are characterized by  new architecture, learning
    methods, predefined parameters and knowledge representation,
    combining the  fuzzy systems capacity of dealing with
    unprecise data and  ANN ability in learning process by
    examples.

    This work proposes a hybrid system, called Neo-Fuzzy-Neuron
    Modified by the Kohonen Network (NFN-MK),
    applied to  the context of function
    identification.  The Neo-Fuzzy-Neuron (NFN) was
    originally proposed by Uchino and Yamakawa
    \cite{uchino94} as an hybrid alternative model
    applied  to real systems.
    The NFN-MK is an  extension of NFN that
    uses a Kohonen Network to generate     initial
    positions of  triangular curves, that
    model the fuzzy neuron.

\section{Description of the Model}


    The NFK-MK model is applied here in the context of function
    approximations, with the objective of adjusting membership
    functions for a neurofuzzy system, using Kohonen
    self-organizing map. The neurofuzzy model used is the
    Neo-Fuzzy-Neuron (NFN) developed by Yamakawa
    \cite{yamakawa94,uchino94}. This model was chosen for its short
    training time, compared to those typical from  multilayer networks
    \cite{caminhas99}.

    The input functions in the system are supposed to be unknown,
    except by samples of
     data. The problem consists in determining a system
    in which the process output $(y_d)$ and the model process
    $(y)$ become close together, according to a given criteria
    \cite{vas99}.

\begin{figure}[H]
  \centering
\includegraphics[height=8cm,width=12cm]{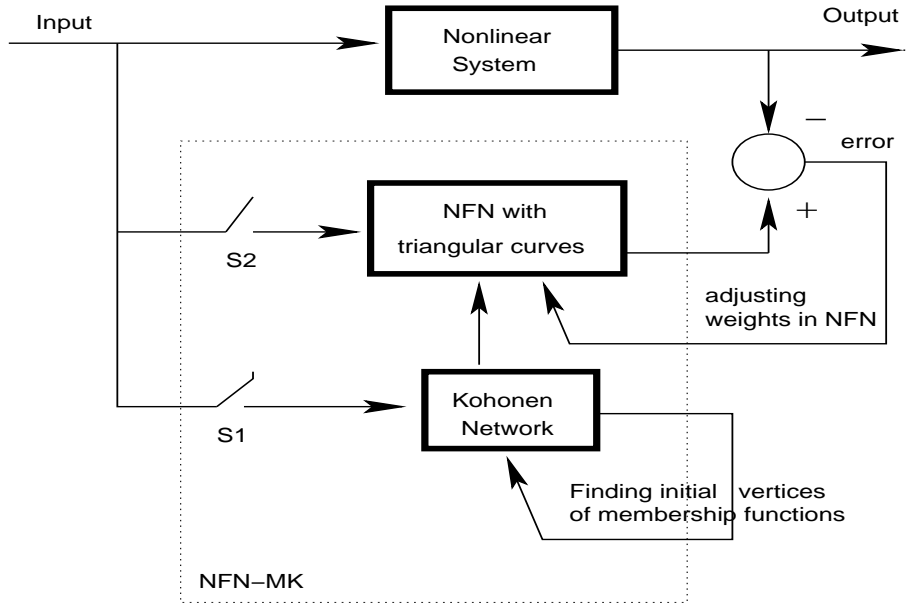}
  \caption{Block diagram for the training phase of NFN-MK}
  \label{2.1}
\end{figure}

    In general we suppose that the process block
    shown in  figure   \ref{2.1} is
    nonlinear. In general this implies in difficulties for the    mathematical
    modeling. The NFN-MK model can be advantageously  used in cases like    this,
    without the necessity of linearizing techniques, unsuitable    for
     some inputs.

     The proposed model  consists in two main blocks
     as seen in figure   \ref{2.1}, included in the
     dashed box. One of them is
     the original Yamakawa's NFN  \cite{yamakawa94,uchino94}.
     It works on triangular curves in its neuron
     fuzzy  model,  used for simplicity.
     In our model the NFN is extended by a
     classical application of Kohonen's network
     \cite{kohonen98}.  As shown in       figure
     \ref{2.1} there are two switches, used for
      the training process.

     When switch $S_2$ is open and $S_1$ closed,
     the Kohonen  network works, looking for the  central
     vertices of the  triangular membership
     functions.  Initially   these values are equally divided
     in seven fuzzy curves, which
      belong to the fuzzy neuron network.
      After this phase,
     the locations of the vertices are updated in
     the NFN block,
     where the base points are found by a new
     training.

     The training on the NFN block occurs
     when  $S_1$ is open and $S_2$ is closed.
     The training proceeds like a backpropagation
     algorithm, finding the weights and base
     points of the triangular curves for each
     neuron.  New points for these triangular
     curves  represent rules such ``{\em
     if-then}'' like original idea of NFN.

    Here seven membership functions
    of triangular type are uniformly distributed in an interval
    $[x_{max},x_{min}]$ of input domain.

     The values of the membership
    functions are based in the experiments of Shaw et al.
    \cite{shaw99}, which show that a change from five to seven
    triangular sets increases the precision in about 15\%. There are
    no significative improvements for greater
    number of sets.

    We note that equidistant membership functions may not be convenient in
    situations where there are concentration of patterns in some
    regions and dispersion in others \cite{shaw99}. One alternative
    is to deal with function non-uniformly distributed. The adjustment
    of the membership functions can be made using a grouping
    algorithm, like Kohonen networks
    \cite{kohonen84,kohonen98}. This is the main reason
    why we are using a very basic Kohonen network,
    that finds  new positions of the vertices of
    triangular curves.  In this method
    the network weights correspond to the values associated to the
    vertices of the triangular curves,
     and the  number of neurons belonging to the processing
    layer corresponds to the number of fuzzy subsets for each
    NFN network input.   The
    winner neuron is the one that corresponds to
    the shortest Euclidean distance from the input
    weight vector \cite{kohonen84,kohonen98}.

\section{Experiment}

A mathematical function that can be used as a benchmark is the
{\em Mexican hat}, defined by

 \begin{equation}
f(x_1,x_2) = \frac{\sin x_1\sin x_2}{x_1 x_2}.
 \label{eq_chapeu}
 \end{equation}

\begin{figure}[H]
  \centering
 \includegraphics[height=7cm,width=11cm]{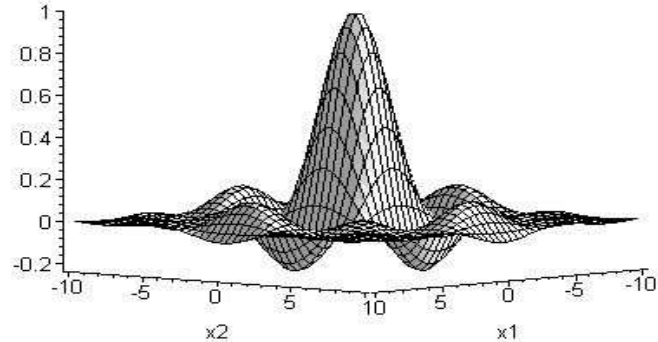}
  \caption{Mexican hat function used as nonlinear system}
  \label{real_mexican}
\end{figure}

This function, shown in  figure \ref{real_mexican}, represents the
nonlinear system to be identified (cf. figure  \ref{2.1}).

Here $x_1 \in [10.0, -10.0] $ and  $x_2 \in [-10.0,10.0]$ are
mapped to $f(x_1,x_2)$  in the interval $[-0.1, 1.0]$.

One reason for  choosing this particular two-variable mathematical
function for testing  our system,  is that the resulting 3D points
can be  easily visualized  in a graphic. The points $(x_1, x_2,
f(x_1,x_2))$ are used for
 training  of Kohonen network and
 also for adjusting the weights in NFN (cf. figure
 \ref{2.1}).

\begin{figure}[H]
  \centering
\includegraphics[height=6cm,width=14cm]{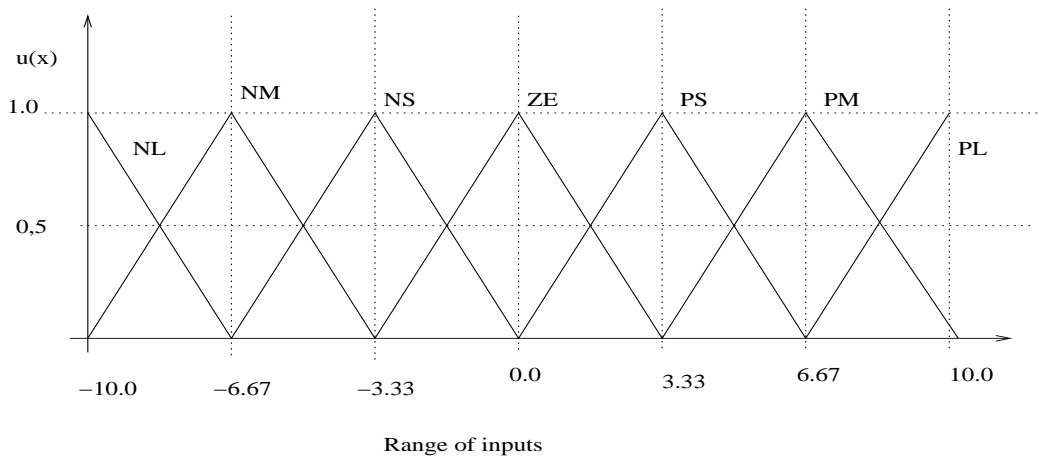}
  \caption{Initial membership functions}
  \label{triangular_1}
\end{figure}

 Initially, the seven membership functions, showed in the figure
\ref{triangular_1}, are equally distributed in the domain. The
vertices as their limits, right and left, are complemented  by
three first columns in table \ref{tabela1}.

The curves in figure  \ref{triangular_1} represent the initial
state of neurons in NFN, which are seven for each input $x_i$
($x_1$ and $x_2$). The semantic values of these curves are
summarized in table \ref{tabela2}, where the notation is the usual
one from fuzzy logic (FL) systems.

The next step consists in  finding  new vertices using a very
basic Kohonen network. These new vertex values are presented  in
the table \ref{tabela1}, (vertex column). This phase is computed
with
 $S_2$ switch  open and $S_1$ closed (cf. figure
 \ref{2.1}).

Once new vertices are found, the triangular curves are redrawn in
according the clustering in each two curves.  These curves are
built to keep  the convexity, in the sense that the summation is 1
for each two curves.
 For example, if $\mu _{ZE}(x) = 0.37$, its
neighbour curve in right side has the complement $\mu _{PS}(x) =
0.63$. By making these adjustments we are following the idea of
clustering
 exhibited  in the  Kohonen network.
Therefore, the new left and right limits for
  these triangular curves are found.

 Since the function
\ref{real_mexican} is symmetric in the two variables, the vertex
positions in the triangular curves are equally distributed. In
this case, the NFN model presents fourteen fuzzy neurons, seven
for each variable ($x_1$ and $x_2$).

\begin{figure}[H]
  \centering
\includegraphics[height=6cm,width=14cm]{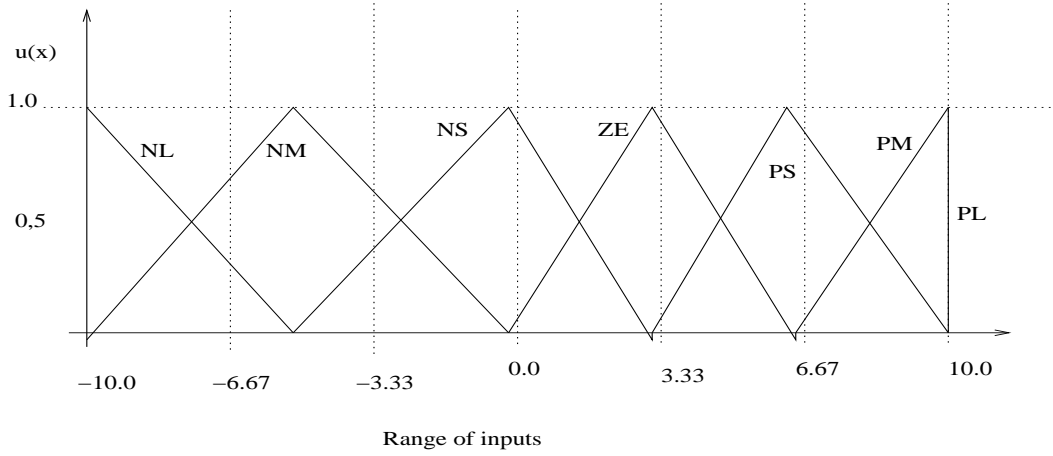}
  \caption{ New membership functions adjusted by NFN-MK}
  \label{triangular_2}
\end{figure}

\begin{figure}[H]
  \centering
  \includegraphics[height=7cm,width=10cm]{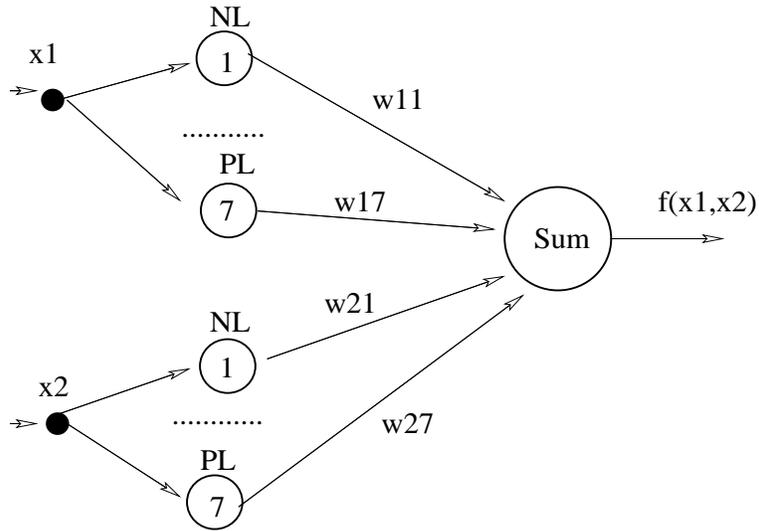}
  \caption{NFN model for two variables ($x_1,x_2$)}
  \label{nfn_classico}
\end{figure}

 By closing switch  $S_2$ and opening $S_1$  (cf. figure
 \ref{2.1}),
 the  adjustment  of weights in the NFN model follows.
 The NFN model is trained with  225 patterns of
 input/output $(x_1,x_2, f(x_1,x_2))$ equally
 distributed, during 10 epochs. The new results
 are presented in  table \ref{tabela1}, and
 the  new values  of the weights in NFN (cf. figure
\ref{nfn_classico}) are  showed in
 last two columns of  table \ref{tabela1}.
 The names of the fuzzy curves   are defined
in table \ref{tabela2}.

\begin{table}[H]
 \centering
\begin{tabular}{|c|c|c|c|c|c|c|c|c|c|}
 \hline
 Fuzzy Curves & \multicolumn{4}{|c|}{Initial Values} &
\multicolumn{5}{|c|}{New Values} \\ \cline{2-10}
    & Left & Vertex & Right &  $w_{1,2}$  & Left & Vertex & Right &
    $\overline{w^1_{final}} $  &  $\overline{w^2_{final}} $ \\ \hline

NL & -10.0 &  -10.0 & -6.67 &  & 10.0 & -10.0 & -3.5 &  0.0715 &  -0.0643 \\
\cline{1-4}\cline{6-10}

NM  & -10.0 & -6.67 & -3.33  &   & -10.0 & -3.5 & -0.2 & -0.1414 &
0.0103 \\ \cline{1-4}\cline{6-10}

NS & -6.67 & -3.33 & 0.0 & & -3.5 & -0.2 & 3.2 & 0.5154 & 0.4973
\\ \cline{1-4}\cline{6-10}

ZE & -3.33 & 0.0 &  3.33 & 1 & -0.2&  3.2 &   6.5 & -0.0824 &
0.0321 \\ \cline{1-4}\cline{6-10}

PS & 0.0 & 3.33 &   6.67 & & 3.2 & 6.5 &  10.0 & 0.0143 &   0.1714
\\ \cline{1-4}\cline{6-10}

PM & 3.33 &  6.67 & 10.0 & & 6.5 & 10.0 & 10.0 &  0.0739 & -0.0463
\\ \cline{1-4}\cline{6-10}

PL & 6.67 &  10.0 & 10.0 & & 10.0 & 10.0 &  10.0 & 0.0 &  0.0358
\\ \hline
\end{tabular}
  \caption{Parameters for the Mexican hat  function }
  \label{tabela1}
\end{table}

\begin{table}[H]
  \centering
\begin{tabular}{ll}
  \hline
NL: & Negative Large \\
NM: & Negative Medium \\
NS: & Negative Small \\
ZE: & Zero \\
PS: & Positive Small \\
PM: & Positive Medium \\
PL: & Positive Large  \\
\hline
\end{tabular}
  \caption{Semantic meanings of fuzzy curves}
  \label{tabela2}

\end{table}

The symbols $\overline{w^1_{final}}$ and $\overline{w^2_{final}}$
denote the average weights of NFN model, considering  each fuzzy
neuron already modified (cf. figure \ref{nfn_classico}). Such
values correspond the following vectors: $\overline{w^1_{final}} =
[w_1^1,w_1^2,w_1^3, ..., w_1^7]$ and $\overline{w^2_{final}} =
[w_2^1,w_2^2,w_2^3, ..., w_2^7]$. Here $w_1^1=w11$, ...,
$w_1^7=w17$, $w_2^1=w21$,..., $w_2^7=w27$, etc.

The resulting sum gives

 \begin{equation}
f(x_1,x_2) = \mu_m(x_1)w_m + \mu_{(m+1)}(x_1)w_{(m+1)} +
\mu_n(x_2)w_n + \mu_{(n+1)}(x_2)w_{(n+1)} \label{fxy}\,.
 \end{equation}

Expression (\ref{fxy}) follows the original idea of Uchino and
Yamakawa \cite{yamakawa94,uchino94}, where the membership
functions are complementary. Thus, the indices $m$ and $m+1$ are
associated to a desfuzzyfication into two complementary curves.
The same idea is applied to $n$ and $n+1$ indices. A numerical
evaluation of the equation (\ref{fxy}), is given  table
\ref{tabela1}.  The  graph in figure \ref{triangular_2} can be
easily computed for any pair of $(x_1,x_2)$.

\section{Evaluation of the Performance}

\begin{figure}
  \centering
  \includegraphics[height=7cm,width=13cm]{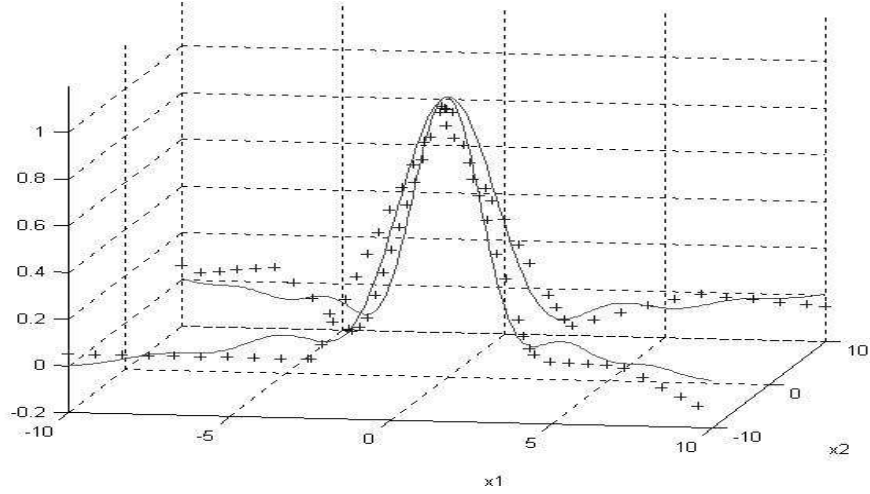}
  \caption{Mexican hat function obtained by NFN-MK}
  \label{hat_simulado}
\end{figure}

The NFN-MK is trained with 225 samples during 10 epochs, so the
internal parameters  become determined (neural weights and
triangular curves of these neurons). After this some inputs are
given. The results of these tests are shown in the figure
\ref{hat_simulado}. Here the points denoted by $+$ represent the
simulation results. The continuous line represents the actual
function.

The values for the training (225 samples and  10 epochs) as well
as the model selected (Mexican hat function), were chosen in order
to compare our results  with those  found elsewhere.
  The parameters chosen
 were provided in these references, except by
 the ANN  model that uses a classical backpropagation algorithm
 with 2 neurons as input layer, 7 other in hidden layer, and 1  in
  output layer. The parameter considered is  the
  number of   mathematical operations necessary  to evaluate a cycle (an epoch),
  for each neuron model in its respective architecture.
 As an simplification, the multiplication, sum and subtraction operations are
 associated to the same value. These results
are shown in table \ref{tabela3}. They show that the NFN-MK
performance  is equivalent to that of other models used to
identify the same curve. Although our system presented a mean
quadratic error (MQE) slightly greater than those obtained by NFHQ
(Neuro-Fuzzy Hierarquic Quadtree) and FSOM (Fuzzy Self-Organized
Map) models, it is much simpler in what concerns mathematical
operations.  On the other hand, the NFN-MK presented a much
smaller MQE than the classical ANN.

\begin{table}[H]
  \centering
\begin{tabular}{|c|c|c|c|}
  \hline
Models  & Number of  Operations ($+$,$-$,$\times $) & Operations
by function& MQE
\\ \hline
NFN-MK &  8 & 2 & 0.0426  \\ \hline NFHQ &  168 & 21 & 0.0150  \\
\hline FSOM &  200 & 101 & 0.0314 \\ \hline NN &  42 & 8 & 0.1037
\\ \hline \hline
\end{tabular}
 \caption{Model comparisions}
  \label{tabela3}
\end{table}


\section{Conclusion}

The objective of this work was to extend  a  NFN model that
allowed the adjustment of membership functions of  triangular type
by the method of Kohonen. The result was  the proposal of the
Neo-Fuzzy-Neuron Modified by a Kohonen Network (NFN-MK). The model
was successfully  tested in function approximation, and its
performance was similar to those obtained by  more complex,
classical neural networks.

The obtained results do not allow inferences   about the
generality of the NFN-MK model, but they seem to indicate that we
have a viable model for system identification. What makes this
model particularly interesting is the  relatively reduced number
of operations and  function calculations involved, implying small
processing times, when compared to other  ANNs
\cite{souza97,rissoli99}.

\bibliographystyle{plain}
\bibliography{kohonen}

\end{document}